\documentclass[letterpaper, 10 pt, conference]{ieeeconf}  % Comment this line out if you need a4paper
\usepackage{hyperref}
\usepackage{graphicx} 
\usepackage{amsmath}
\usepackage{amssymb}
\usepackage{multirow}
\usepackage{xcolor}

\IEEEoverridecommandlockouts                              % This command is only needed if 
\title{\LARGE \bf

ROG-Grasp: Root-Oriented Geometry for Robotic Grasping and Placement
}

\begin{document}

% \author{Anonymous Authors}
\author{
Zijian An$^{1}$, Augustus Sroka$^{1}$, Ran Yang$^{2}$, Bill Cai$^{1}$, Satoru Eto$^{1}$, Brian Poon$^{1}$, Kelvin Cai$^{1}$, Shijie Geng$^{3}$, \\ Feng Liu$^{1}$, Yiming Feng$^{2}$, and Lifeng Zhou$^{1\star}$
% \thanks{: }%
\thanks{$^{\star}$ Corresponding author} %
\thanks{$^{1}$ Department of Electrical and Computer Engineering, Drexel University, 3141 Chestnut St, Philadelphia, PA, 19104, USA} %
\thanks{$^{2}$ Virginia Seafood Agricultural Research and Extension Center, and Department of Biological Systems Engineering, Virginia Tech, 15 Rudd Ln, Hampton, VA 23669, USA} %
\thanks{$^{3}$ Amazon Store Foundation AI (SFAI), 12 W 39th St, New York, NY 10018, USA}
}

\maketitle
\thispagestyle{empty}
\pagestyle{empty}

%%%%%%%%%%%%%%%%%%%%%%%%%%%%%%%%%%%%%%%%%%%%%%%%%%%%%%%%%%%%%%%%%%%%%%%%%%%%%%%%
\begin{abstract}
Orientation-aware manipulation is essential in post-harvest agricultural processing, where produce must be grasped and placed in consistent configurations. 
This paper presents ROG-Grasp, a geometry-based robotic grasping and placement framework that estimates the produce orientation from root surface geometry using RGB-D perception. 
A YOLO-based root detector and point cloud plane fitting are used to infer the root normal, enabling stable grasp pose generation and orientation-constrained Cartesian motion planning. 
Experiments on tomatoes and onions demonstrate high success rates and stable execution time in both isolated and cluttered scenarios. 
Compared with vision–language–action (VLA) policies, the proposed method achieves more reliable and accurate grasp completion with faster execution. These results highlight the effectiveness of geometry-driven perception for practical orientation-controlled manipulation tasks. \href{https://youtu.be/Ir2UtGODdMo}{A video} of our paper is available online. 
\end{abstract}

%%%%%%%%%%%%%%%%%%%%%%%%%%%%%%%%%%%%%%%%%%%%%%%%%%%%%%%%%%%%%%%%%%%%%%%%%%%%%%%%
\section{Introduction}

Automation in agricultural production and post-harvest processing has attracted increasing attention due to rising labor costs and the demand for higher productivity. Robotic systems have therefore been widely investigated for tasks such as produce harvesting, sorting, and packaging in modern agricultural supply chains. Vision-based perception, robotic manipulation, and end-effector design are considered key enabling technologies for intelligent agricultural robots \cite{zhang2024automatic, huang2025review, tan2025review}. 
In particular, vision-guided robotic manipulators have been extensively studied for automated fruit harvesting and handling, where perception systems detect target produce and robotic arms perform grasping and transfer operations \cite{zhang2025review, tang2020recognition, droukas2023survey}. For example, strawberry harvesting robots equipped with machine vision and robotic grippers have been developed to autonomously detect and pick ripe strawberries in greenhouse environments \cite{xiong2020autonomous, xiong2018design}. Similarly, robotic harvesting systems for apples integrate vision modules with industrial manipulators and vacuum grippers to locate and pick fruits in orchard settings \cite{krakhmalev2022robotic}. Recent work has also explored robotic manipulation strategies for tomatoes using vision-based detection and adaptive grippers designed for handling soft agricultural products \cite{ansari2025novel}. These studies demonstrate the feasibility of robotic perception and grasping for agricultural products.

\begin{figure*}
    \centering
    \includegraphics[width=1.0\linewidth]{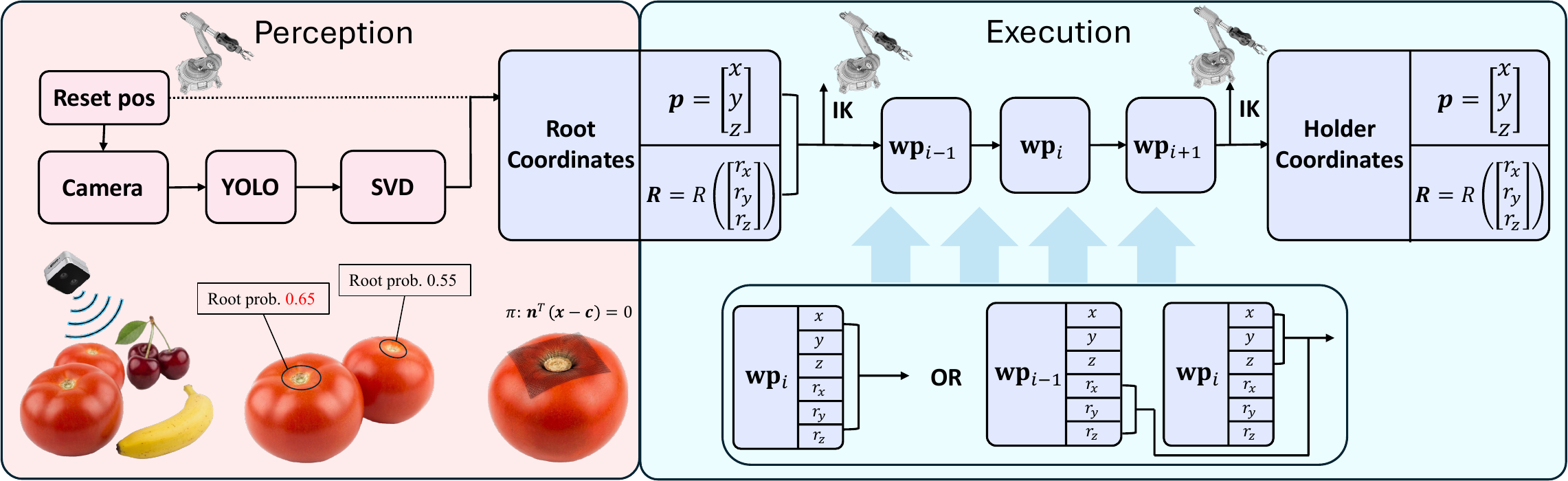}
    \caption{Overview of the proposed ROG-Grasp framework. 
    The pipeline consists of two main modules: vision perception and motion execution. In the perception stage, a YOLO-based detector identifies the root region and a plane-fitting procedure estimates the produce orientation from the extracted point cloud. In the execution stage, the robot follows a sequence of intermediate waypoints to safely perform grasping and upright placement while avoiding collisions.}
    \label{arch_fig}
\end{figure*}

However, beyond harvesting, post-harvest processing and packaging introduce additional requirements on manipulation accuracy and object orientation. In modern packhouses, fruits and vegetables must be sorted, transferred, and packed into trays or containers before transportation and retail \cite{mulholland2024adoption, dewi2020fruit}. Robotic systems have therefore been explored for automated fruit packing and handling in industrial environments, where robots detect produce and place them into packaging boxes or containers \cite{zhang2025multi}. In some packaging lines, robotic manipulators are capable of arranging fruits in containers while maintaining consistent orientation or layout for efficient packing and improved product presentation \cite{oladeleunderstanding}. Such orientation-aware placement is particularly relevant for high-value produce, where visually organized arrangements can enhance protection and commercial appearance.

Despite these advances, most existing agricultural robotic systems primarily focus on fruit detection and grasping, while the problem of pose-controlled placement remains relatively underexplored \cite{pan2023panoptic, beldek2025multi}. Many harvesting robots simply pick fruits and deposit them into containers without explicitly controlling the final orientation of the object \cite{wang2022geometry, chu2024high}. However, in practical agricultural and packaging scenarios, produce often need to be placed in specific orientations to satisfy the requirements of downstream processing, inspection, or packaging operations. For example, certain produce such as garlic, onions, and hawthorn may need to be positioned consistently during packaging or processing workflows \cite{chi2013direction, chen2022design, fang2024determination, li2020design}. Existing grasp planning methods frequently rely on generic grasp detection or bounding box information, which makes it difficult to infer the intrinsic orientation of irregular agricultural products.

Recent studies have explored the use of vision–language–action (VLA) models to address robotic manipulation problems, demonstrating promising levels of generalization and decision-making capability in complex engineering scenarios \cite{an2025claw, yang2025seqvla,zhong2025dexgraspvla,zhu2025vla}. Motivated by these advances, we also investigated the use of VLA-based policies for orientation-aware grasping of agricultural products. However, our preliminary experiments indicate that such models often struggle to reliably identify the root region of produce. As a result,
% purely learning-based
end-to-end learning
approaches may have difficulty achieving the orientation estimation accuracy required for stable grasping and controlled placement.

To address this limitation, this paper proposes a robotic grasping method based on root surface geometry for orientation-aware manipulation of agricultural products, as shown in Figure~\ref{arch_fig}. The key observation is that many agricultural products, such as tomatoes, kiwifruits, garlic, and onions, contain a root or stem surface that exhibits approximately planar geometry with relatively small curvature. By detecting the root region and extracting its point cloud from RGB-D observations, a plane fitting method can be applied to estimate the produce orientation from the root surface. The estimated root normal provides a reliable cue for determining object orientation and generating a stable grasp pose for robotic manipulation. The proposed method enables accurate orientation-aware grasping and placement of produce, providing a practical solution for robotic agricultural manipulation tasks that require controlled placement orientation.

The primary contributions of this paper are as follows:
\begin{itemize}
\item A geometry-based method that uses root surface information to estimate the orientation of agricultural products from RGB-D observations, enabling orientation-aware robotic grasping.
\item A vision-guided grasp pose generation strategy that integrates YOLO root segmentation with 3D point cloud plane fitting to infer object orientation and compute stable grasp poses.
\item Experimental validation on multiple agricultural products, including tomato and onion, demonstrating reliable orientation-aware grasping and placement, and improved robustness compared with VLA-based policies in root localization and pose-controlled manipulation.

\end{itemize}

\section{Related work}
\textbf{YOLO}. Deep learning–based object detection has become a standard perception approach in robotics. 
Among existing detectors, the YOLO (You Only Look Once) family is widely adopted due to its ability to achieve real-time performance with high detection accuracy using single-stage inference \cite{jiang2022review}. 
YOLO-based models have been extensively applied in agricultural vision tasks such as fruit detection, produce monitoring, and robotic harvesting, where fast and reliable perception is essential for manipulation in complex outdoor environments \cite{sapkota2026comprehensive}.

In this work, a YOLO-based segmentation model is employed to detect the root region of agricultural products. 
The detected region is then converted into a 3D point cloud from RGB-D observations, enabling geometric orientation estimation and grasp pose generation.
% \section{Problem Formulation}

\textbf{Robot Motion Control}. Industrial manipulators typically support both joint-space and Cartesian motion commands. 
Joint-space motion specifies target joint configurations directly, enabling fast repositioning. 
Cartesian motion commands specify a desired robot flange pose in task space $SE(3)$, which is commonly parameterized by position $\mathbf{p}\in\mathbb{R}^3$ and orientation $\mathbf{R}\in\mathbb{R}^{3\times 3}$, requiring inverse kinematics (IK) to compute the corresponding joint trajectories. 
For a typical 6-DoF manipulator, the IK problem may admit multiple feasible solutions. 
In our implementation, up to eight candidate joint configurations can be obtained for a given Cartesian pose. 
To ensure smooth and continuous motion execution, the solution closest to the current joint configuration is selected.

In this work, all robot motions are executed using Cartesian motion commands specified by position $\mathbf{p}$ and orientation $\mathbf{R}$, as illustrated in Figure~\ref{arch_fig}. This design choice is motivated by the fact that agricultural manipulation tasks are naturally defined in Euclidean space. In these scenarios, the spatial position and desired orientation of the target object (e.g., a produce or a holder) can be directly obtained from perception modules, whereas the corresponding joint configurations of the manipulator are not directly available.
Position $\mathbf{p}$ and orientation $\mathbf{R}$ can be compactly represented by a homogeneous transformation matrix
\begin{equation}
\mathbf{T} =
\begin{bmatrix}
\mathbf{R} & \mathbf{p} \\
\mathbf{0}^\top & 1
\end{bmatrix}\in SE(3),
\label{t}
\end{equation}
which provides a unified representation of the robot flange pose, including both rotation and translation. 
The robot controller computes the corresponding joint configurations based on the desired flange pose in task space. 
However, the grasping operation requires accurate positioning of the gripper rather than the robot flange. 
Therefore, a coordinate transformation between the gripper frame and the robot flange frame must be considered.
This transformation is formulated using homogeneous transformation matrices in $SE(3)$, which provide a unified representation of rigid-body motion including both rotation and translation between coordinate frames. 
The detailed derivation of the transformation and its integration into grasp pose generation will be presented in Section \ref{motion_planning}.

\begin{figure}
    \centering
    \includegraphics[width=\linewidth]{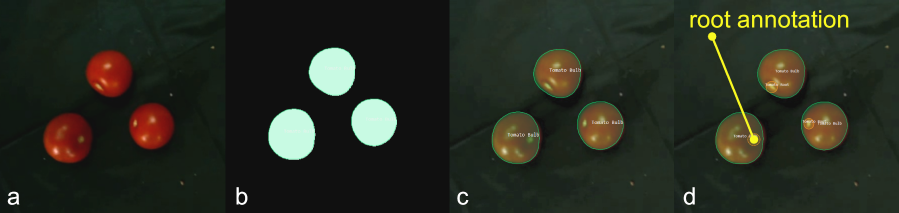}
    \caption{YOLO dataset annotation process.}
    \label{yolo}
\end{figure}
\section{Approach}
% A YOLO-based detector is employed to identify the root region of the crop from RGB images. Based on the detected region and the corresponding depth data, the system estimates the crop orientation and generates a stable grasping pose, enabling the robot to place the crop in the desired upright configuration.
% The overall pipeline of the proposed approach is illustrated in Figure~ \ref{arch_fig}.
The overall workflow of the proposed system is illustrated in Figure~\ref{arch_fig}. 
Starting from a predefined observation pose, the RGB-D camera captures the scene and the visual perception module detects the root region of the target produce. 
Based on the detected root point cloud, the system estimates the root orientation and computes the corresponding grasp pose. 
The computed pose is then converted into Cartesian motion commands and executed through inverse kinematics to generate a sequence of intermediate waypoints for safe manipulation.

After grasping the produce, the robot adjusts the placement orientation and transports the object to the predefined holder location for upright placement. 
The entire pipeline can therefore be divided into two main components: visual perception for root detection and motion planning and control for grasp pose generation and execution. 
The details of these two components are described in the following subsections.
\begin{figure*}
    \centering
    \includegraphics[width=0.85\linewidth]{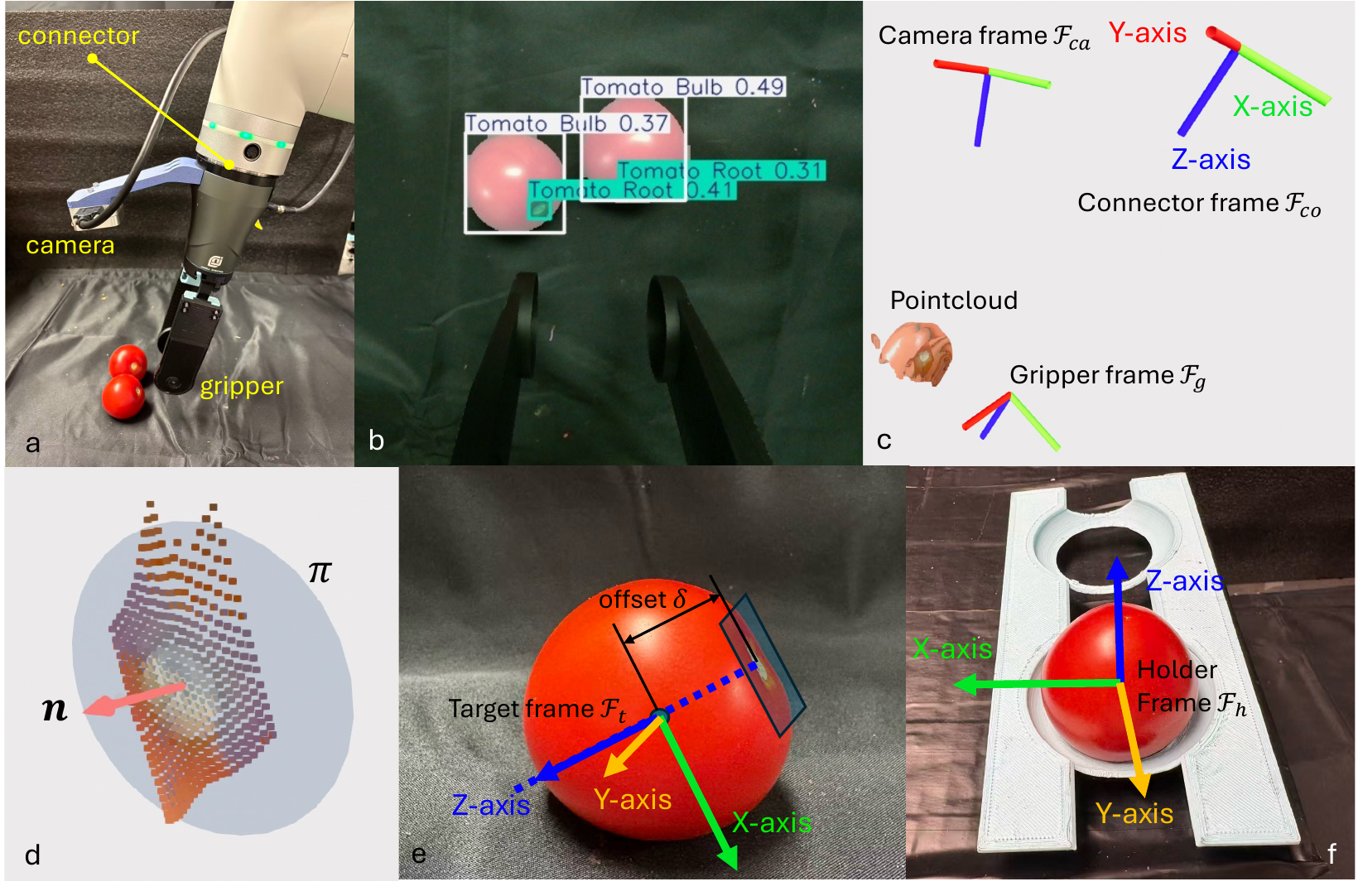}
    \caption{Key components of geometry-based orientation estimation and pose generation. }
    \label{approach_fig}
\end{figure*}
\subsection{Visual Perception for Root Detection}
\label{visual_perception}
% \textbf{System Overview.} 
Our system comprises a robotic manipulator equipped with an RGB-D camera. The camera observes the workspace and provides both color and depth information for perception.

\textbf{Training Dataset and YOLO Fine-Tuning.} To construct a training dataset, a video of the target produce is recorded against a contrasting backdrop, capturing a representative subset of camera viewpoints. Individual frames are extracted, and polygon annotations for the produce body are automatically generated by computing a foreground mask against the contrasting backdrop, as illustrated in Figure~\ref{yolo}. Root annotations must be manually created for each root instance. This semi-automatic pipeline substantially reduces the manual effort required for dataset construction. The annotated dataset is further augmented through lighting and orientation transformations to improve model generalization. A YOLO-26 segmentation model is then fine-tuned on this dataset to produce body and root segmentation.

\textbf{Root Detection and Point Cloud Extraction.} At inference time, the YOLO-based detector identifies the produce body and any visible root region, as illustrated in Figure~\ref{approach_fig}(b). Among the detected produce body candidates, the region with the highest detection confidence is selected for further processing. 
% \au{If a root segmentation intersects the selected produce body instance, the system proceeds with orientation estimation.\lz{dont understand}} 
Using the corresponding depth image, the pixels within the detected body and root region are projected into 3D space to generate a point cloud representation of the root surface in the camera coordinate frame. Since the depth sensor provides metric depth values, each pixel can be back-projected into 3D using the camera intrinsic parameters. The resulting point cloud is then transformed from the camera coordinate frame to the robot base frame with camera-to-base transformation.

\subsection{Motion Planning and Control}
\label{motion_planning}

As illustrated in Figure~\ref{approach_fig}(c), the spatial relationships among the key coordinate frames introduced in Figure~\ref{approach_fig}(a) are shown. The base coordinate frame is denoted by $\mathcal{F}_{\text{base}}$ and is defined at the base of the robotic manipulator. The connector frame $\mathcal{F}_{\text{co}}$, camera frame $\mathcal{F}_{\text{ca}}$, and gripper frame $\mathcal{F}_{\text{g}}$ are attached to the connector, camera, and gripper, respectively.

The RGB-D camera captures the point cloud in the camera frame $\mathcal{F}_{\text{ca}}$.
To enable motion planning and grasp pose generation, each 3D point is transformed into the base frame using the calibrated camera-to-base homogeneous transformation
\begin{equation}
\tilde{\mathbf{p}}_{\text{base}} =
\mathbf{T}_{\text{base}}^{\text{ca}} \tilde{\mathbf{p}}_{\text{ca}},
\label{homo_p}
\end{equation}
where $\tilde{\mathbf{p}}_{\text{ca}}, \tilde{\mathbf{p}}_{\text{base}} \in \mathbb{R}^4$ denote homogeneous point coordinates in camera frame and base frame, respectively.
The transformation matrix $\mathbf{T}_{\text{base}}^{\text{ca}}$ follows the standard homogeneous representation defined in Equation~\eqref{t}.

\textbf{Root Pose Estimation}. Root produce such as garlic, onions, and hawthorn typically exhibit a root surface that approximates a circular planar region with relatively small curvature. Therefore, the root orientation can be estimated by fitting a plane to the extracted root point cloud.
To estimate the orientation of the tomato root surface, a least-squares plane fitting method is applied to the extracted root point cloud. 
Let $\{\mathbf{p}_i\}_{i=1}^{N}$ denote the set of 3D points belonging to the detected root region of the tomato. 
Each point $\mathbf{p}_i = (x_i, y_i, z_i)^\top$ represents the Euclidean coordinates corresponding to the homogeneous point $\tilde{\mathbf{p}}_{\text{base}} = (x_i, y_i, z_i, 1)^\top$ obtained by Equation \ref{homo_p}.
The centroid of the root point cloud is computed as
\[
\mathbf{c} = \frac{1}{N}\sum_{i=1}^{N} \mathbf{p}_i .
\]
Each point is then centered with respect to the centroid,
\[
\mathbf{q}_i = \mathbf{p}_i - \mathbf{c}.
\]
The centered point matrix is decomposed using singular value decomposition (SVD). 
The direction corresponding to the smallest singular value represents the direction with the minimum spatial variance of the root point cloud and therefore defines the normal vector of the fitted plane,
\[
\mathbf{n} \in \mathbb{R}^3.
\]
The root plane can be represented in vector form as
\[
\mathbf{\pi}:\mathbf{n}^\top(\mathbf{x}-\mathbf{c}) = 0,
\]
The normal vector $\mathbf{n}$ describes the orientation of the tomato root surface $\pi$ and is therefore used to estimate the orientation of the tomato.

\textbf{Grasp and Placement Pose Generation.} 
% After estimating the root orientation, the grasping pose is generated based on the centroid of the root point cloud 
The grasp pose is parameterized by an orientation obtained from the fitted root surface normal $\mathbf{n}$ and a position defined as the centroid of the root point cloud.
Let $\mathbf{c}$ denote the centroid of the root point cloud. 
As illustrated in Figure~\ref{approach_fig}(e), the grasping position is obtained by applying an offset along the normal direction,
\[
\mathbf{p}_{\text{target}} = \mathbf{c} + \delta \mathbf{n},
\]
where $\delta$ is a predefined offset distance. This offset places the grasp point slightly above the root surface, allowing the manipulator to approach the target produce more precisely.
%from a stable direction 

A target coordinate frame $\mathcal{F}_t$ is then defined at $\mathbf{p}_{\text{target}}$, whose $z$-axis is aligned with the normal vector $\mathbf{n}$. 
This target frame represents the desired grasp pose of the produce. 
By aligning the gripper frame with the target frame, i.e., $\mathcal{F}_{g} = \mathcal{F}_{\text{t}}$, the manipulator moves the gripper to the computed grasp position with the appropriate orientation, enabling reliable picking and placement of the produce in the desired upright configuration.

Finally, the placement pose is defined by the holder frame, as illustrated in Figure~\ref{approach_fig}(f). 
The holder frame $\mathcal{F}_{h}$ is fixed in the robot workspace and specifies the desired upright placement configuration. 
To place the produce into the holder, the robot is commanded such that the gripper frame coincides with the holder frame, i.e., $\mathcal{F}_{g} = \mathcal{F}_{h}$. 
This alignment guides the manipulator to move the grasped produce to the holder and release it in the desired upright configuration.
 
The pose of the holder frame $\mathcal{F}_{h}$ is obtained through a teaching procedure. 
Specifically, the robot is manually guided until the gripper reaches the desired placement configuration inside the holder. 
At this configuration, the homogeneous transformation of the connector frame with respect to the base frame, denoted by $\mathbf{T}_{\text{base}}^{\text{co}*}$, is recorded from the robot controller.
Since the rigid transformation between the gripper frame and the connector frame, $\mathbf{T}_{\text{co}}^{\text{g}}$, is fixed, the corresponding gripper placement position can be computed using homogeneous coordinates as
\begin{equation*}
\tilde{\mathbf{p}}_{g}^{*} = \mathbf{T}_{\text{base}}^{\text{co}*}\mathbf{T}_{\text{co}}^{\text{g}} \tilde{\mathbf{p}}_{0},
\label{eq_g}
\end{equation*}
where $\tilde{\mathbf{p}}_{0}=(0,0,0,1)^\top$ denotes the origin of the gripper frame.

The resulting point $\mathbf{p}_{g}^{*}$ defines the desired placement location. 
During execution, the desired placement orientation of the gripper is first determined by the motion planning stage, which specifies the connector rotation matrix $\mathbf{R}_{\text{base}}^{\text{co}}$. 
Since the rigid transformation between the connector frame and the gripper frame is fixed, the position of the connector can be expressed as
\begin{equation}
\begin{aligned}
\mathbf{p}_{\text{base}}^{\text{co}}
&=
\mathbf{p}_{g}^{*} -\mathbf{R}_{\text{base}}^{\text{co}} \mathbf{t}_{\text{co}}^{g}\\
&=\mathbf{T}_{\text{base}}^{\text{co}*}\mathbf{T}_{\text{co}}^{\text{g}} \tilde{\mathbf{p}}_{0} - \mathbf{R}_{\text{base}}^{\text{co}} \mathbf{t}_{\text{co}}^{g},
\label{eq_p}
\end{aligned}
\end{equation}
where $\mathbf{t}_{\text{co}}^{g}$ denotes the constant translational offset from the connector frame to the gripper frame.
The resulting Cartesian pose $\mathbf{T}_{\text{base}}^{\text{co}}$, formed by combining the planned orientation and the computed connector position, is then sent to the robot controller for execution.
\section{Experiments}
\subsection{Execution Pipeline}
The experimental setup consists of a Fairino FR5 robotic manipulator equipped with a Jodell RG52-050 gripper. 
The perception system is provided by the RGB-D camera  $\text{D405}$, as shown in Figure \ref{approach_fig}(a).
The experiments were conducted using two types of root produce, namely onions and tomatoes. 
The task objective is to grasp the detected produce and place it upright into a predefined holder. 
The holder position is fixed in the robot workspace and defines the desired placement pose.

\begin{figure}
    \centering
    \includegraphics[width=0.85\linewidth]{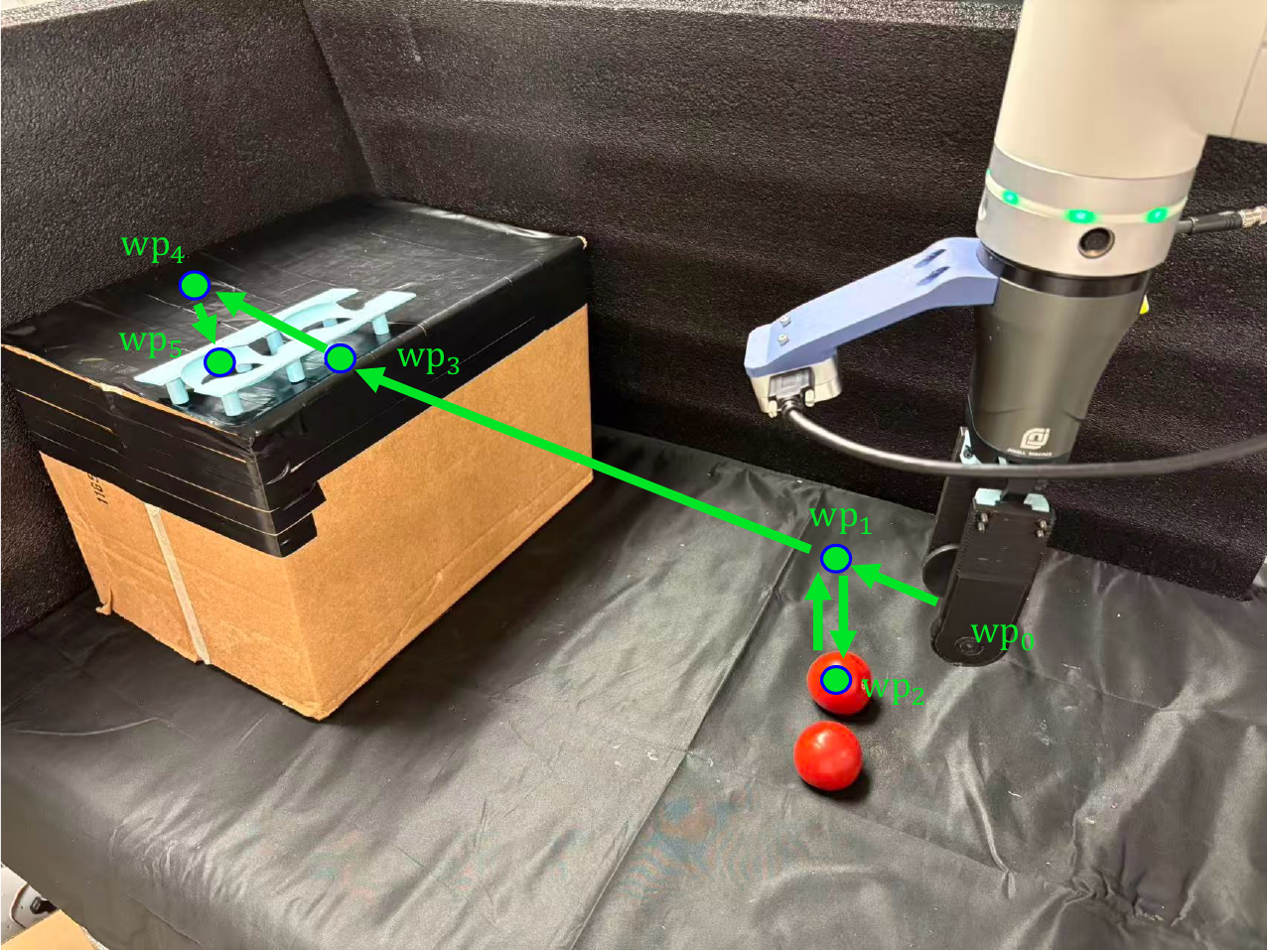}
    \caption{Waypoint-based execution pipeline for orientation-aware grasping and placement. 
% The robot moves through a sequence of predefined and dynamically generated waypoints to approach the crop, perform grasping, adjust placement orientation, and transport the object to the fixed holder location while ensuring collision-free motion.
}
    \label{workflow_1}
\end{figure}

Considering the workspace limitations of the robotic manipulator and safety requirements during motion execution, a set of intermediate waypoints is predefined. 
The complete execution workflow is illustrated in Figure~\ref{workflow_1}, where the robot follows six sequential waypoints denoted as $\texttt{wp}_0$–$\texttt{wp}_5$. 
The waypoint $\texttt{wp}_0$ represents the initial observation pose, from which the camera captures the scene and the YOLO detector identifies the root region of the produce. After the perception stage, the robot moves through the subsequent waypoints to approach the produce, perform grasping, and finally place the produce into the holder.
\begin{figure*}
    \centering
    \includegraphics[width=\linewidth]{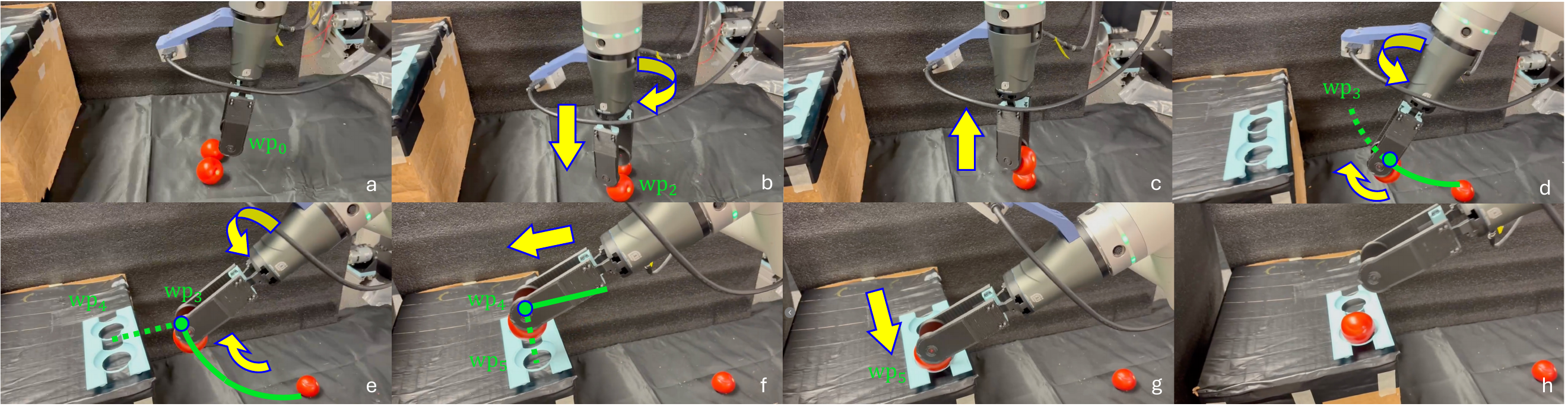}
    \caption{Sequential execution of orientation-aware grasping and placement for a tomato.}
    \label{workflow_2}
\end{figure*}

\begin{figure*}
    \centering
    \includegraphics[width=\linewidth]{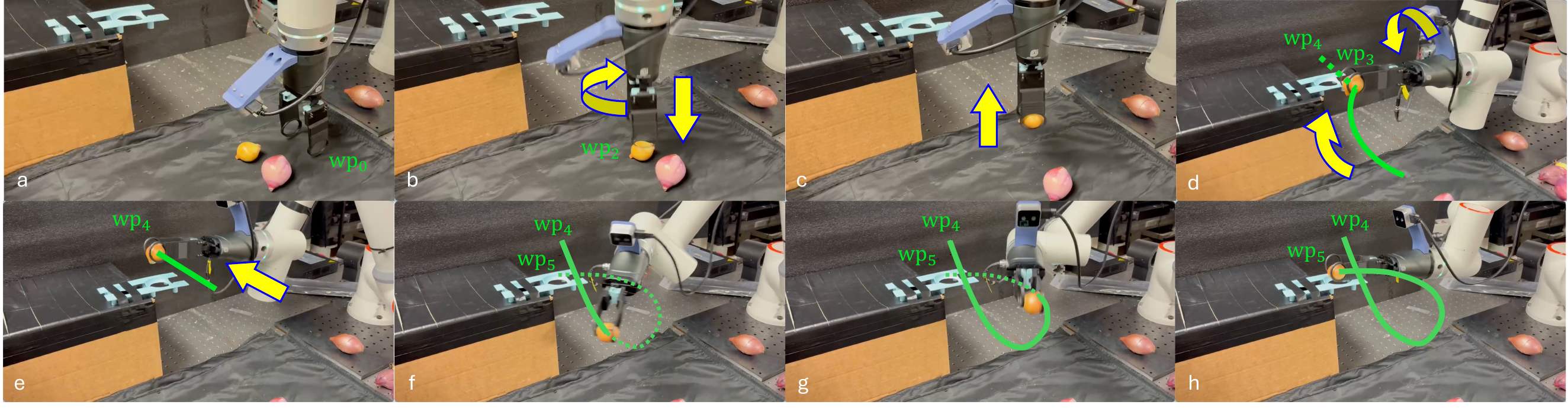}
    \caption{Sequential execution of orientation-aware grasping and placement for an onion. Notably, in subfigures (f)--(h), the robot moves from waypoint $\mathrm{wp}_4$ to $\mathrm{wp}_5$, with $\mathrm{wp}_5$ located directly below $\mathrm{wp}_4$. While a straight downward motion may seem sufficient in Cartesian space, maintaining the same end-effector orientation would lead to a joint-limit violation at joint~4. As a result, the inverse-kinematics solution satisfying the joint constraints yields the observed arc-shaped trajectory instead of a straight vertical path.}
    \label{workflow_3}
\end{figure*}

Figure~\ref{workflow_2} and \ref{workflow_3} illustrate a complete manipulation process for picking and placing a tomato and an onion, respectively. The manipulation process for tomato and onion is similar. For simplicity, the following description focuses on the tomato example.
Figure~\ref{workflow_2}(a)–(h) shows the key stages of the execution.
In Figure~\ref{workflow_2}(a), the robot starts from the initial observation pose, where the scene is captured, and the YOLO detector identifies the root region of the produce. 
Based on the detected root location, the robot first moves to an approach waypoint $\texttt{wp}_1$ located above the target object and then descends vertically to $\texttt{wp}_2$ to approach the produce from the top. 
This two-stage approach avoids potential collisions with surrounding objects and prevents disturbing the target produce during motion. 
Figure~\ref{workflow_2}(b) shows the approaching process toward $\texttt{wp}_2$, during which the robot adjusts Joint 6 to align the gripper orientation with the estimated grasp direction.

After the produce is grasped, the robot first retreats upward back to $\texttt{wp}_1$, as shown in Figure~\ref{workflow_2}(c), to avoid interference with other objects on the table. 
The robot then moves to waypoint $\texttt{wp}_3$. 
During this motion, the placement orientation is adjusted to ensure that the produce will be placed upright, i.e., with the root facing downward. 
Figure~\ref{workflow_2}(d) and (e) illustrate this adjustment process, where the orientation is mainly controlled through Joint 4 and Joint 6.

Waypoints $\texttt{wp}_1$, $\texttt{wp}_2$, and $\texttt{wp}_3$ are dynamically generated according to the detected produce position and initial orientation, whereas $\texttt{wp}_4$ and $\texttt{wp}_5$ are fixed as they correspond to the predefined holder location. 
From $\texttt{wp}_3$ onward, the connector orientation $\mathbf{R}_{\text{base}}^{\text{co}}$ remains unchanged while only the Cartesian position $\mathbf{p}_{\text{base}}^{\text{co}}$ is updated, ensuring that the desired placement orientation is preserved.
Figure~\ref{workflow_2}(f) and (g) show the robot reaching $\texttt{wp}_4$ and the final placement pose $\texttt{wp}_5$, respectively. 
The corresponding connector Cartesian pose $\mathbf{T}_{\text{base}}^{\text{co}}$ is computed using Equations~\eqref{eq_p}. 
After releasing the produce, the robot retreats upward to avoid disturbing the placed produce and then returns to the initial pose, as illustrated in Figure~\ref{workflow_2}(h).

\subsection{VLA-Based Manipulation Evaluation}
\begin{table}[t]
\centering
\caption{Quantitative Comparison Between ROG-Grasp and VLA-Based Policies}
\label{tab:performance}
\begin{tabular}{|c|c|cc|cc|}
\hline
\multicolumn{2}{|c|}{} & \multicolumn{2}{c|}{ROG-Grasp} & \multicolumn{2}{c|}{VLA} \\ \cline{3-6}
\multicolumn{2}{|c|}{} & Tomato & Onion & Tomato & Onion \\ \hline

\multirow{2}{*}{Single} 
& Time (s) & 8.3 & 7.2 & 20.7 & 22.3 \\ \cline{2-6}
& Succ. Rate (\%) & 85 & 90 & 40 & 35 \\ \hline

\multirow{2}{*}{Multi} 
& Time (s) & 8.2 & 7.5 & 25.4 & 26.6 \\ \cline{2-6}
& Succ. Rate (\%) & 80 & 80 & 10 & 5 \\ \hline

\end{tabular}
\end{table}

To evaluate learning-based approaches for orientation-aware grasping, we trained a VLA policy using the $\pi_0$ model \cite{black2410pi0}. 
The training dataset consists of 50 teleoperated episodes recorded from two RGB-D cameras: a wrist-mounted $\text{D405}$ providing the same viewpoint as in our geometry-based method, and a fixed overhead $\text{D455}$ observing the entire workspace. 
Due to the observed performance differences between isolated and cluttered environments, separate policies were trained for single-object and multi-object scenarios.

Table~\ref{tab:performance} compares the proposed ROG-Grasp method with the VLA policy in terms of execution time and success rate. 
A trial is considered successful if the produce is placed into the holder with its root facing downward within an angular tolerance of $\pm 20^\circ$. 
ROG-Grasp achieves consistently higher success rates and significantly shorter execution times across all settings.

Failures of ROG-Grasp mainly occur during the descent from $\texttt{wp}_1$ to $\texttt{wp}_2$, where unintended contact may disturb the object pose. 
As this phase is executed in open loop, the system cannot perform real-time corrections. 
In contrast, the VLA policy operates in a closed loop by generating action chunks of length 50 based on visual feedback. However, this leads to a substantially longer execution time.  Moreover, the policy often fails to infer appropriate placement orientations and struggles in cluttered scenes, where it may oscillate between nearby objects without establishing a stable grasp.

\section{Conclusion and Future Work}

This paper presented ROG-Grasp, a geometry-based approach for orientation-aware robotic grasping and placement of agricultural products. 
Experimental results demonstrate that the proposed method achieves stable and fast execution time and high success rates across different produce types and task settings. 
Compared with VLA-based policies, ROG-Grasp provides more reliable grasp completion and consistent placement orientation, particularly in cluttered environments. 
These findings highlight the effectiveness and practical robustness of geometry-driven orientation estimation for agricultural manipulation tasks that require controlled object placement.

Future work will focus on extending the proposed framework toward more adaptive and generalized manipulation capabilities. 
One important direction is to incorporate closed-loop visual feedback during the grasping phase to improve robustness against unexpected object motion and contact disturbances. 
Another promising direction is to integrate geometry-based orientation estimation with learning-based policies to enable improved perception and decision-making in highly cluttered or unstructured environments. 
Furthermore, future studies will investigate manipulating a wider range of agricultural products with more complex shapes and extend the system to multi-object task planning scenarios.

%%%%%%%%%%%%%%%%%%%%%%%%%%%%%%%%%%%%%%%%%%%%%%%%%%%%%%%%%%%%%%%%%%%%%%%%%%%%%%%%

\bibliographystyle{IEEEtran}
\bibliography{refs}

@article{an2025claw,
  title={CLAW: A Vision-Language-Action Framework for Weight-Aware Robotic Grasping},
  author={An, Zijian and Yang, Ran and Feng, Yiming and Zhou, Lifeng},
  journal={arXiv preprint arXiv:2509.14143},
  year={2025}
}

@article{yang2025seqvla,
  title={SeqVLA: Sequential Task Execution for Long-Horizon Manipulation with Completion-Aware Vision-Language-Action Model},
  author={Yang, Ran and An, Zijian and Zhou, Lifeng and Feng, Yiming},
  journal={arXiv preprint arXiv:2509.14138},
  year={2025}
}

@article{zhang2024automatic,
  title={Automatic fruit picking technology: A comprehensive review of research advances},
  author={Zhang, Jun and Kang, Ningbo and Qu, Qianjin and Zhou, Lianghuan and Zhang, Hongbo},
  journal={Artificial Intelligence Review},
  volume={57},
  number={3},
  pages={54},
  year={2024},
  publisher={Springer}
}

@article{huang2025review,
  title={A review of visual perception technology for intelligent fruit harvesting robots},
  author={Huang, Yikun and Xu, Shuyan and Chen, Hao and Li, Gang and Dong, Heng and Yu, Jie and Zhang, Xi and Chen, Riqing},
  journal={Frontiers in Plant Science},
  volume={16},
  pages={1646871},
  year={2025},
  publisher={Frontiers Media SA}
}

@article{tan2025review,
  title={A review of research on fruit and vegetable picking robots based on deep learning},
  author={Tan, Yarong and Liu, Xin and Zhang, Jinmeng and Wang, Yigang and Hu, Yanxiang},
  journal={Sensors},
  volume={25},
  number={12},
  pages={3677},
  year={2025},
  publisher={MDPI}
}

@article{zhang2025review,
  title={A review on the recent developments in vision-based apple-harvesting robots for recognizing fruit and picking pose},
  author={Zhang, Yanqiang and Li, Na and Zhang, Lijie and Lin, Jianfeng and Gao, Xiao and Chen, Guangyi},
  journal={Computers and Electronics in Agriculture},
  volume={231},
  pages={109968},
  year={2025},
  publisher={Elsevier}
}

@article{tang2020recognition,
  title={Recognition and localization methods for vision-based fruit picking robots: A review},
  author={Tang, Yunchao and Chen, Mingyou and Wang, Chenglin and Luo, Lufeng and Li, Jinhui and Lian, Guoping and Zou, Xiangjun},
  journal={Frontiers in plant science},
  volume={11},
  pages={510},
  year={2020},
  publisher={Frontiers Media SA}
}

@article{droukas2023survey,
  title={A survey of robotic harvesting systems and enabling technologies},
  author={Droukas, Leonidas and Doulgeri, Zoe and Tsakiridis, Nikolaos L and Triantafyllou, Dimitra and Kleitsiotis, Ioannis and Mariolis, Ioannis and Giakoumis, Dimitrios and Tzovaras, Dimitrios and Kateris, Dimitrios and Bochtis, Dionysis},
  journal={Journal of Intelligent \& Robotic Systems},
  volume={107},
  number={2},
  pages={21},
  year={2023},
  publisher={Springer}
}

@article{xiong2020autonomous,
  title={An autonomous strawberry-harvesting robot: Design, development, integration, and field evaluation},
  author={Xiong, Ya and Ge, Yuanyue and Grimstad, Lars and From, P{\aa}l J},
  journal={Journal of Field Robotics},
  volume={37},
  number={2},
  pages={202--224},
  year={2020},
  publisher={Wiley Online Library}
}

@inproceedings{xiong2018design,
  title={Design and evaluation of a novel cable-driven gripper with perception capabilities for strawberry picking robots},
  author={Xiong, Ya and From, Pal Johan and Isler, Volkan},
  booktitle={2018 IEEE international conference on robotics and automation (ICRA)},
  pages={7384--7391},
  year={2018},
  organization={IEEE}
}

@article{krakhmalev2022robotic,
  title={Robotic complex for harvesting apple crops},
  author={Krakhmalev, Oleg and Gataullin, Sergey and Boltachev, Eldar and Korchagin, Sergey and Blagoveshchensky, Ivan and Liang, Kang},
  journal={Robotics},
  volume={11},
  number={4},
  pages={77},
  year={2022},
  publisher={MDPI}
}

@article{ansari2025novel,
  title={A novel approach to tomato harvesting using a hybrid gripper with semantic segmentation and keypoint detection},
  author={Ansari, Shahid and Gohil, Mahendra Kumar and Maeda, Yusuke and Bhattacharya, Bishakh},
  journal={arXiv preprint arXiv:2512.03684},
  year={2025}
}

@article{mulholland2024adoption,
  title={The adoption of robotics in pack houses for fresh produce handling},
  author={Mulholland, Barry J and Panesar, Pardeep S and Johnson, Philip H},
  journal={The Journal of Horticultural Science and Biotechnology},
  volume={99},
  number={1},
  pages={9--19},
  year={2024},
  publisher={Taylor \& Francis}
}

@article{oladeleunderstanding,
  title={Understanding Decision Making for Automation in Packhouse and Human Capital Requirement},
  author={Oladele, Olabisi Bisi}
}

@article{dewi2020fruit,
  title={Fruit sorting robot based on color and size for an agricultural product packaging system},
  author={Dewi, Tresna and Risma, Pola and Oktarina, Yurni},
  journal={Bulletin of Electrical Engineering and Informatics},
  volume={9},
  number={4},
  pages={1438--1445},
  year={2020}
}

@article{zhang2025multi,
  title={Multi-arm robotic system and strategy for the automatic packaging of apples},
  author={Zhang, Yizhi and Chen, Liping and Li, Xin and Li, Qicheng and Li, Jiangbo},
  journal={Artificial Intelligence in Agriculture},
  year={2025},
  publisher={Elsevier}
}

@inproceedings{pan2023panoptic,
  title={Panoptic mapping with fruit completion and pose estimation for horticultural robots},
  author={Pan, Yue and Magistri, Federico and L{\"a}be, Thomas and Marks, Elias and Smitt, Claus and McCool, Chris and Behley, Jens and Stachniss, Cyrill},
  booktitle={2023 IEEE/RSJ International Conference on Intelligent Robots and Systems (IROS)},
  pages={4226--4233},
  year={2023},
  organization={IEEE}
}

@inproceedings{beldek2025multi,
  title={Multi-vision-based Picking Point Localisation of Target Fruit for Harvesting Robots},
  author={Beldek, C and Dunn, A and Cunningham, J and Sariyildiz, Emre and Phung, SL and Alici, G},
  booktitle={2025 IEEE International Conference on Mechatronics (ICM)},
  pages={1--6},
  year={2025},
  organization={IEEE}
}

@article{wang2022geometry,
  title={Geometry-aware fruit grasping estimation for robotic harvesting in apple orchards},
  author={Wang, Xing and Kang, Hanwen and Zhou, Hongyu and Au, Wesley and Chen, Chao},
  journal={Computers and Electronics in Agriculture},
  volume={193},
  pages={106716},
  year={2022},
  publisher={Elsevier}
}

@article{chu2024high,
  title={High-precision fruit localization using active laser-camera scanning: Robust laser line extraction for 2D-3D transformation},
  author={Chu, Pengyu and Li, Zhaojian and Zhang, Kaixiang and Lammers, Kyle and Lu, Renfu},
  journal={Smart Agricultural Technology},
  volume={7},
  pages={100391},
  year={2024},
  publisher={Elsevier}
}

@article{chi2013direction,
  title={Direction identification system of garlic clove based on machine vision},
  author={Chi, Gao and Hui, Gao},
  journal={TELKOMNIKA Indonesian Journal of Electrical Engineering},
  volume={11},
  number={5},
  pages={2323--2329},
  year={2013}
}

@article{chen2022design,
  title={Design and experiment of a garlic orientation and orderly conveying device based on machine vision},
  author={Chen, Jianneng and Yu, Chennan and Yao, Kun and Zhou, Yun and Zhou, Binsong},
  journal={Agriculture},
  volume={12},
  number={8},
  pages={1077},
  year={2022},
  publisher={MDPI}
}

@article{fang2024determination,
  title={Determination of garlic clove orientation based on capacitive sensing technology},
  author={Fang, Lizhi and Zhou, Kai and Li, Tianhua and Hou, Jialin and Li, Yuhua},
  journal={Computers and Electronics in Agriculture},
  volume={219},
  pages={108827},
  year={2024},
  publisher={Elsevier}
}

@article{li2020design,
  title={Design and experiment of adjustment device based on machine vision for garlic clove direction},
  author={Li, Yuhua and Wu, Yanqiang and Li, Tianhua and Niu, Ziru and Hou, Jialin},
  journal={Computers and Electronics in Agriculture},
  volume={174},
  pages={105513},
  year={2020},
  publisher={Elsevier}
}

@article{jiang2022review,
  title={A Review of Yolo algorithm developments},
  author={Jiang, Peiyuan and Ergu, Daji and Liu, Fangyao and Cai, Ying and Ma, Bo},
  journal={Procedia computer science},
  volume={199},
  pages={1066--1073},
  year={2022},
  publisher={Elsevier}
}

@article{sapkota2026comprehensive,
  title={Comprehensive performance evaluation of yolov12, yolo11, yolov10, yolov9 and yolov8 on detecting and counting fruitlet in complex orchard environments},
  author={Sapkota, Ranjan and Meng, Zhichao and Churuvija, Martin and Du, Xiaoqiang and Ma, Zenghong and Karkee, Manoj},
  journal={Agriculture Communications},
  pages={100125},
  year={2026},
  publisher={Elsevier}
}

@article{zhong2025dexgraspvla,
  title={Dexgraspvla: A vision-language-action framework towards general dexterous grasping},
  author={Zhong, Yifan and Huang, Xuchuan and Li, Ruochong and Zhang, Ceyao and Chen, Zhang and Guan, Tianrui and Zeng, Fanlian and Lui, Ka Num and Ye, Yuyao and Liang, Yitao and others},
  journal={arXiv preprint arXiv:2502.20900},
  year={2025}
}

@article{zhu2025vla,
  title={VLA-Grasp: a vision-language-action modeling with cross-modality fusion for task-oriented grasping},
  author={Zhu, Jianwei and Sun, Xueying and Zhang, Qiang and Liu, Mingmin},
  journal={Complex \& Intelligent Systems},
  volume={11},
  number={6},
  pages={272},
  year={2025},
  publisher={Springer}
}

@article{black2410pi0,
  title={$\pi_0$: A vision-language-action flow model for general robot control. CoRR, abs/2410.24164, 2024. doi: 10.48550},
  author={Black, Kevin and Brown, Noah and Driess, Danny and Esmail, Adnan and Equi, Michael and Finn, Chelsea and Fusai, Niccolo and Groom, Lachy and Hausman, Karol and Ichter, Brian and others},
  journal={arXiv preprint ARXIV.2410.24164}
}

\end{document}